\documentclass[runningheads]{llncs}
\usepackage[T1]{fontenc}
\usepackage{booktabs}
\usepackage{graphicx}
\usepackage{multirow}
\usepackage{amsmath,amssymb}
\usepackage{graphicx}
\usepackage{hyperref}
\usepackage{makecell}
\usepackage{chngpage}
\usepackage{color}
\usepackage[misc]{ifsym}
\usepackage{mathtools}
\usepackage{wrapfig}
\renewcommand{\hat}{\widehat}
\newcommand{\wt}{\widetilde}
\newcommand{\wh}{\widehat}

\DeclareMathOperator{\R}{{\mathbb R}}

\DeclareMathOperator{\A}{{\mathcal A}}
\DeclareMathOperator{\cS}{{\mathcal S}}

\DeclareMathOperator*{\disp}{\mathrm{\Delta}}

\newcommand{\eg}{{\em e.g.,~}}           
\newcommand{\ie}{{\em i.e.,~}}

\newcommand{\XL}{\textcolor{blue}}

\newcommand{\revision}{\textcolor{black}} 

\newcommand{\ours}{\textbf{\texttt{FCRO}}}

\begin{document}

\def\thefootnote{*}\footnotetext{These authors contributed equally to this work.}
\title{On Fairness of Medical Image Classification \\ with Multiple Sensitive Attributes via \\ Learning Orthogonal Representations}
%
\titlerunning{On Fairness of Image Classification with Multi-Sensitive Attributes}
\author{Wenlong~Deng$^*$\inst{1}\orcidID{0009-0002-0545-6384} \and Yuan~Zhong$^*$\inst{2}\orcidID{0009-0004-4909-8763} \and Qi~Dou\inst{2}\orcidID{0000-0002-3416-9950} \and Xiaoxiao~Li\inst{1}\orcidID{0000-0002-8833-0244}}
%
\authorrunning{W. Deng et al.}
%
\institute{Department of Electrical and Computer Engineering,\\ The University of British Columbia, Vancouver, BC, Canada 
 \email{\{dwenlong,xiaoxiao.li\}@ece.ubc.ca}
\and
Department of Computer Science and Engineering,\\ The Chinese University of Hong Kong, Hong Kong, China
 \email{\{yzhong22,qdou\}@cse.cuhk.edu.hk}}

%
\maketitle              
\begin{abstract}
Mitigating the discrimination of machine learning models has gained increasing attention in medical image analysis. However, rare works focus on fair treatments for patients with multiple sensitive demographic attributes, which is a crucial yet challenging problem for real-world clinical applications. 
In this paper, we propose a novel method for fair representation learning with respect to multi-sensitive attributes. We pursue the independence between target and multi-sensitive representations by achieving orthogonality in the representation space. 
Concretely, we enforce the column space orthogonality by keeping target information on the complement of a low-rank sensitive space.
Furthermore, in the row space, we encourage feature dimensions between target and sensitive representations to be orthogonal. 
The effectiveness of the proposed method is demonstrated with extensive experiments on the CheXpert dataset. To our best knowledge, this is the first work to mitigate unfairness with respect to multiple sensitive attributes in the field of medical imaging. The code is available at \href{https://github.com/ubc-tea/FCRO-Fair-Classification-Orthogonal-Representation}{https://github.com/ubc-tea/FCRO-Fair-Classification-Orthogonal-Representation}. 

\end{abstract}
 \vspace*{-0.5\baselineskip}


%
%
%
\section{Introduction}



With the increasing application of artificial intelligence systems for medical image diagnosis, it is notably important to ensure fairness of image classification models and investigate concealed model biases that are to-be-encountered in complex real-world situations.
Unfortunately, sensitive attributes (e.g., race and gender) accompanied by medical images are prone to be inherently encoded by machine learning models \cite{glocker2021algorithmic}, and affect the model's discrimination property~\cite{zhang2022improving}. 
Recently, fair representation learning has shown great potential as it acts as a group parity bottleneck that mitigates discrimination when generalized to downstream tasks. Existing methods \cite{adeli2021representation,dullerud2022fairness,sarhan2020fairness,shui2022learning} have studied the parity between privileged and unprivileged groups upon just a single sensitive attribute, but neglecting the flexibility with respect to multiple sensitive attributes, in which the conjunctions of unprivileged attributes might deteriorate discrimination. This is a crucial yet challenging problem hindering the applicability of machine learning models, especially for medical image classification where patients always have many demographic attributes. 
\begin{figure}[!tp]
    \centering
    \includegraphics[width=0.92\linewidth]{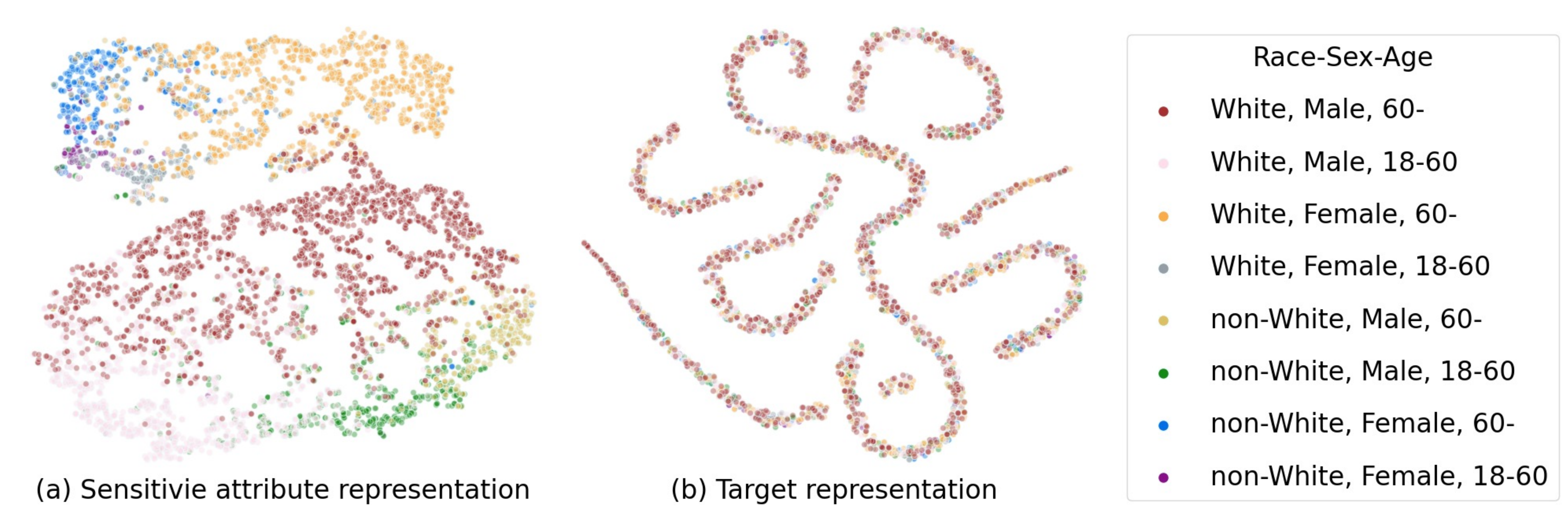}
    \caption{A t-SNE \cite{van2008visualizing} visualization of  (a) sensitive attribute and (b) target representations learned from our proposed methods \ours{} on the CheXpert dataset \cite{irvin2019chexpert}. Sensitive embeddings capture subgroups' variance. We claim \ours{} enforces fair classification on the target task by learning orthogonal target representations that are invariant over different demographic attribute combinations.
    }
    \label{fig:tsne}
\end{figure}

To date, it is still challenging to effectively learn target-related representations which are both fair and flexible to multiple sensitive attributes, regardless of some promising investigations recently. For instance, adversarial methods \cite{adeli2021representation,madras2018learning} produce robust representations by formulating a min-max game between an encoder that learns class-related representation and an adversary that removes sensitive information from it. 
Disentanglement-based methods \cite{dullerud2022fairness} achieve separation by minimizing the mutual information between target and sensitive attribute representations. These methods typically gain efficacy by means of carefully designed objectives. To extend them to the multi-attribute setting, additional loss functions have to be explored, which should handle gradient conflict or interference. 
Methods using variational autoencoder~\cite{creager2019flexibly} decompose the latent distributions of target and sensitive and penalize their correlation for disentanglement. However, aligning the distribution of the sensitive attributes is difficult or even intractable given the complex combination of multiple factors. 
Besides, there are some fairness methods based on causal inference \cite{madras2019fairness} or bi-level optimization \cite{shui2022learning}, which also learn debiased while multi-attributes inflexible representations. Recently, disentanglement is vigorously interpreted as the orthogonality of a decomposed target-sensitive latent representation pair by \cite{sarhan2020fairness}, where they predefine a pair of orthogonal subspaces for target and sensitive attribute representations. 
However, in the multi-sensitive attributes setting, the dimension of the target space would be continuously compressed and how to solve it is still an open problem.
In this paper, we propose a new method to achieve \underline{F}airness via \underline{C}olumn-\underline{R}ow space \underline{O}rthogonality (dubbed \ours{}) by learning fair representations with respect to multiple sensitive attributes for medical image classification.
\ours{} considers multi-sensitive attributes by encoding them into a unified attribute representation. It achieves a best trade-off for fairness and data utility (see illustrations in Fig.~\ref{fig:tsne}) via orthogonality in both column and row spaces. Our contributions are summarized as follows: (1) We tackle the practical and challenging problem of fairness given multiple sensitive attributes for medical image classification. To the best of our knowledge, this is the first work to study fairness with respect to multi-sensitive attributes in the field of medical imaging. (2) We relax the independence of target and sensitive attribute representations by orthogonality which can be achieved by our proposed novel column and row losses.
(3) We conduct extensive experiments on the CheXpert \cite{irvin2019chexpert} dataset with over 80,000 chest X-rays. \ours{} achieves a superior fairness-utility trade-off over state-of-the-art methods regarding multiple sensitive attributes race, sex, and age.
\section{Methodology}

\subsection{Problem Formulation}
\textbf{Notations.}
We consider group fairness in this work. \revision{While keeping the model utility for the privileged group,} group fairness articulates the equality of some statistics like predictive rate between certain groups. Considering a binary classification problem with column vector inputs $x \in \mathcal{X}$, labels $y \in \mathcal{Y} = \{0,1\}$. \revision{\textit{Multi-sensitive attributes} $a \in \mathcal{A}$ is vector of $m$ attributes sampled from the conjunction of binary attributes}, \ie Cartesian product of sensitive attributes $\mathcal{A} = \prod_{i\in[m]}\footnote{$[m]=\{0, 1, .. , m\}$}A_i$, where $A_i\in\{0,1\}$ is the $i$-th sensitive attribute. Our training data consist of tuples $\mathcal{D}=\{(x,y,a)\}$. We denote the classification model $f(x)=h_T(\phi_T(x))$ that predicts a class label $\hat{y}$ given $x$, where $\phi_T:\mathcal{X} \mapsto \R^d$ is a feature encoder for target embeddings, and $h_T:\R^d \mapsto \R$ is a scoring function. Similarly, we consider a sensitive attribute model $g(x)=\{h_{A_1}(\phi_A(x)),...,h_{A_m}(\phi_A(x))\}$ that predicts sensitive attributes associated with input $x$. Given the number of samples $n$, the input data representation is $X=[x_1, \dots, x_n]$ and we denote the feature representation $Z_T=\phi_T(X)$, $Z_A=\phi_A(X)$ $\in \R^{d \times n}$. 
\begin{figure}[!t]
    \centering    \includegraphics[width=0.9\linewidth]{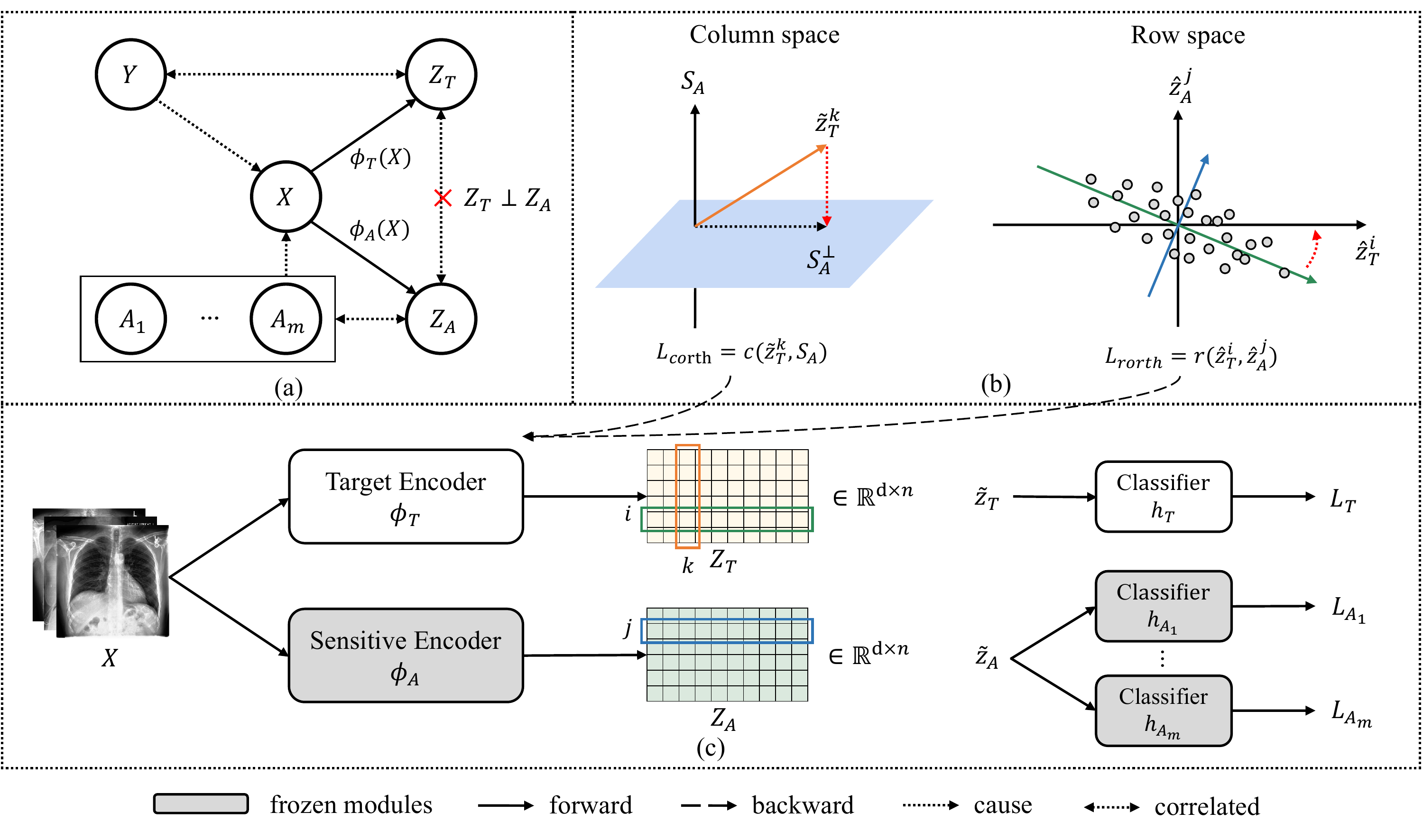}
    \caption{Overview of our proposed method \ours. (a) The graphical model of orthogonal representation learning for fair medical image classification with multiple sensitive attributes. (b) The novel column-row space orthogonality. In the column space, we encourage the target model to learn representations in the complement of a low-rank sensitive space. In the row space, we enforce each row vector (feature dimension) of the target and sensitive attribute representations to be orthogonal to each other (c) The overall training pipeline. We use a pre-trained multi-sensitive branch, and propagate orthogonal gradients to target encoder $\phi_T$. }
    \label{fig:method}
     \vspace*{-0.5\baselineskip}
\end{figure}

\noindent\textbf{Fairness with multiple sensitive attributes.}
A classifier predicts $y$ given an input $x$ by estimating the posterior probability $p(y|x)$. When inputs that are affected by their associated attributes (\ie $\{A_1,\dots,A_m\} \rightarrow X$) are fed into the network, the posterior probability is written as $p(y|x,a)$. Since biased information from $\A$ is encoded, this can lead to an unfair prediction by the classifier. For example, in the diagnosis of a disease with sensitive attributes age, sex, and race, a biased classifier will result in $p(\hat{y} | A = {\rm male},{\rm old},{\rm black}) \neq p(\hat{y}|A = \rm{female}, \rm{young}, \rm{white})$. In this work, we focus on equalized odds (ED), which is a commonly used and crucial criterion of fair classification in the medical domain \cite{xu2022survey}. In our case, ED regarding multiple sensitive attributes can be formulated as follows: 
\begin{align} \label{eq:cjme}
    P(\hat{Y}=y|A=\pi_1,Y=y) = P(\hat{Y}=y|A=\pi_2,Y=y), \ \forall \pi_1,\pi_2 \in \mathcal{A},y\in\mathcal{Y}.
\end{align}
A classifier satisfies ED with respect to group $A$ and the target $Y$ if the model prediction $\hat{Y}$ and $A$ are conditionally independent given $Y$, \ie $\hat{Y} \perp A | Y$.

\noindent\textbf{Fair representation.}
To enforce our aforementioned conditions, we follow \cite{sarhan2020fairness} and introduce target embedding $z_T$ and multi-attribute embedding $z_{A_i}$ that is generated from $x$. As in the causal structure graph for the classifier depicted in Fig.~\ref{fig:method}~(a), our objective is to find a fair model that maximizes the log-likelihood function of the distribution $p(y,a|x)$, where:
\begin{align}
    p(y,a|x) = \frac{p(y|x,a)p(x|a)p(a)}{p(x)} & = p(y|x)p(a|x) \label{eq:loglike}\\
    & = p(y|z_T)p(z_T|x)\prod_{i\in[m]}p(a_i|z_{A_i})p(z_{A_i}|x), \label{eq:minde}
\end{align}
and we call $z_T$ \textit{fair representation} for the target task (\eg disease diagnosis). To this end, we aim to maximize Eq.~\eqref{eq:minde} with the conditional independence constraint to train a fair classifier. 
It is noteworthy that in the multi-sensitive attributes setting, forcing $z_T$ to be independent on all $z_{A_i}, \forall i\in[m]$ is challenging and even intractable when $m$ is large.
Therefore, we propose to encode multi-sensitive attributes into a \textit{single} compact encoding $z_A$ that is still predictive for classifying attributes (\ie $z_A \rightarrow \{a_1, \dots, a_m\}$). Then we can rewrite Eq.~\eqref{eq:minde} as likelihood with the independence constraint on $z_T$ and $z_A$: 
\begin{align}\label{eq:inde}
    p(y,a|x) = p(y|z_T)p(z_T|x)p(a|z_A)p(z_A|x).
\end{align}
However, maximizing Eq.~\eqref{eq:inde} brings two technical questions:\\
\textbf{Q1:} How to satisfy the independence constraint for $z_T$ and $z_A$?\\
\textbf{A1:} 
We relax independence by enforcing orthogonality \revision{in a finite vector space}. Different from predefined orthogonal space in \cite{sarhan2020fairness}, we enforce orthogonality in both column spaces (Sec.~\ref{sec:space}) and row spaces (Sec.~\ref{sec:row}) of $Z_T$ and $Z_A$. \\
\textbf{Q2:} How to estimate $p(y|z_T)$, $p(z_T|x)$, $p(a|z_A)$, $p(z_A|x)$?\\
\textbf{A2:} We train two convolutional encoders $z_T = \phi_T(x)$ and $z_A = \phi_A(x)$ to approximate $p(z_T|x)$ and $p(z_A|x)$ respectively; And we train two multi-layer perception classifiers $ y=h_T(z_T)$ and $a = h_A(z_A) $ to approximate $p(y|z_T)$ and $p(a|z_A)$ respectively (Sec.~\ref{sec:overall}). 

\subsection{Column Space Orthogonality}\label{sec:space}
First, we focus on the column space of the target and the sensitive attribute representations. Column space orthogonality aims to learn target representations $Z_T$ 
that fulfill the following \textbf{two aims}: 1) have the least projection onto the sensitive space $\mathcal{S}_A$ and 2) preserve the representation power to predict $Y$.

Denote the target representation $Z_T=[\wt{z}_{T}^1, \wt{z}_{T}^2,\dots, \wt{z}_{T}^n]$ and the sensitive attribute representation $Z_A=[\wt{z}_{A}^1, \wt{z}_{A}^2,\dots, \wt{z}_{A}^n]$, where $\wh{z}^i \in \R^{d\times 1}$ is a column vector for $i \in [n]$, we represent the column space for $Z_T$ and $Z_A$ as $\cS_T = {\rm span}(Z_T)$ and $\cS_A= {\rm span} (Z_A)$ respectively.
\textbf{Aim 1} can be achieved by forcing $\cS_T = \cS_A^{\perp}$. Although both $\wt{z}_{T},\wt{z}_{A}\in \R^d$, their coordinates may not be aligned as they are generated from two separate encoders.
As a result, if $d \ll \infty$, then there is no straightforward way to achieve $\cS_T \perp \cS_A$ by directly constraining $\wt{z}^i_{T},\wt{z}^j_{A}$ (\eg forcing $(\wt{z}^i_{T})^\top \wt{z}^j_{A}=0$). \textbf{Aim 2} can be achieved by seeking a low-rank representation $\wt{\cS}_A$ for $\cS_A$, whose rank is $k$ such that $k \ll d$, because we have ${\rm rank}(\cS_T)+{\rm rank}(\cS_A) = d$ if $\cS_T = \cS_A^{\perp}$ holds. Then $\cS_A^\perp$ would be a high-dimensional space with sufficient representation power for target embeddings. This is especially important when we face multiple sensitive attributes, as the total size of the space is $d$, and increasing the number of sensitive attributes would limit the capacity of $\cS_T$ to learn predictive $\wt{z}_{T}$. 
To this end, we first propose to find the low rank sensitive attribute representation space $\wt{\cS}_A$, and then encourage $Z_T$ to be in $\wt{\cS}_A$'s complement $\wt{\cS}_A^\perp$. 

\noindent\textbf{Construct low-rank multi-sensitive space.}\label{sec:space_acc} We apply Singular Value Decomposition (SVD) on $Z_A = U_A \Sigma_A V_A$ to construct the low-rank space $\wt{\mathcal{S}}_A$, where $U_A,V_A\in \R^{d\times d}$ are orthogonal matrices with left and right singular vectors $u_A^i \in \R^{d}$ and $v^i_A \in \R^{n}$ respectively. And $\Sigma_A \in \R^{d\times n}$ is a diagonal matrix with descending non-negative singular values $\{\delta^i_A\}_{i=1}^{min\{n,d\}}$. Then we extract the most important $k$ left singular vectors to construct $\wt{\mathcal{S}}_A = [u^1_A, ... , u^k_A]$, where $k$ controls how much sensitive information to be captured in $\wt{\cS}_A$. It is notable that $\wt{\mathcal{S}}_A$ is agnostic to the number of sensitive attributes because they share the same $Z_A$. For situations that can not get the whole dataset at once, we follow~\cite{lin2022trgp} to select the most important bases from both bases of old iterations and newly constructed ones, thus providing an accumulative low-rank space construction variant to update $\wt{\cS}_A$ iteratively. As we do not observe significant performance differences between these two variants (Fig.~\ref{fig:curve}~(a)), we use and refer to the first one in this paper if there is no special clarification.

\noindent\textbf{Column orthogonal loss.} With the low-rank space $\wt{\mathcal{S}}_A$ for multiple sensitive attributes, we encourage $\phi_T$ to learn representations in its complement $\wt{\mathcal{S}}_A^\perp$. Notice that $\wt{\mathcal{S}}_A^\perp$ can also be interpreted as the kernel of the projection onto $\wt{\mathcal{S}}_A$, \ie $\wt{\mathcal{S}}_A^\perp=\text{Ker}(\text{proj}_{\wt{\mathcal{S}}_A}\wt{z}_T)$.
Therefore, we achieve column orthogonal loss by minimizing the projection of $Z_T$ to $\wt{\mathcal{S}}_A$,  which can be defined as:
\begin{equation}
    \begin{aligned}
    L_{corth} &= c(Z_T,\wt{\mathcal{S}}_A) = \sum^n_{i=1}\frac{\left\|\wt{\mathcal{S}}_A^\top \wt{z}_{T}^i\right\|_2^2}{\left\|\wt{z}_T^i\right\|_2^2}.
\end{aligned}
\end{equation}
As $\wt{\mathcal{S}}_A$ is a low-rank space, $\wt{\mathcal{S}}_A^\perp$ will have abundant freedom for $\phi_T$ to extract target information, thus reserving predictive ability. 
\subsection{Row Space Orthogonality}\label{sec:row} 
Then, we study the row space of target and sensitive attribute representations. Row space orthogonality aims to learn target representations $Z_T$ that have the least projection onto the sensitive row space $\hat{\cS}_A$. In other words, we want to ensure orthogonality on each feature dimension between $Z_T$ and $Z_A$. Denote target representation $Z_T=[\hat{z}_{T}^1; \hat{z}_{T}^2;\dots; \hat{z}_{T}^d]$ and sensitive attribute representation $Z_A=[\hat{z}_{A}^1; \hat{z}_{A}^2;\dots; \hat{z}_{A}^d]$, where $\hat{z}^i \in \R^{1\times n}$ is a row vector for $i \in [d]$. 
We represent row space for target representations and sensitive attribute representations as $\hat{\cS}_T = {\rm span}(Z_T^{\top})$ and $\hat{\cS}_A= {\rm span} (Z_A^{\top})$ correspondingly. 
As the coordinates (\ie the index of samples) of $\hat{z}_{A}$ and $\hat{z}_{T}$ are aligned, forcing $\hat{\cS}_T \perp \hat{\cS}_A$ can be directly applied by pushing $\hat{z}_T^i$ and $\hat{z}_A^j$ to be orthogonal for arbitrary $i,j\in[d]$. 

Unlike column space, the orthogonality here won't affect the utility, since the row vector $\hat{z}_{T}$ is not directly correlated to the target $y$. To be specific, we let pair-wise row vectors $Z_T=[\hat{z}_{T}^1, \hat{z}_{T}^2,\dots, \hat{z}_{T}^d]$ and $Z_A=[\hat{z}_{A}^1, \hat{z}_{A}^2,\dots, \hat{z}_{A}^d]$ have a zero inner dot-product. Then for any $i,j \in [d]$, we try to minimize $<\wh{z}_{T}^i, \wh{z}_{A}^j >$. Here we slightly modify the orthogonality by extra subtracting the mean vector $\mu_A$ and $\mu_T$ from $Z_A$ and $Z_T$ respectively, where $\mu = \mathbb{E}_{i\in[d]}\hat{z}^i \in \mathbb{R}^{1\times n}$. Then orthogonality loss will naturally be integrated into a covariance loss:

\begin{equation}
    \begin{aligned}
L_{rorth}=r(Z_T,Z_A)= \frac{1}{d^2}\sum^d_{i=1}\sum^d_{j=1} \left[(\hat{z}^i_{T}-\mu_{T})(\hat{z}^j_{A}-\mu_{A})^{\top}\right]^2.
\end{aligned}
\end{equation} 
In this way, the resulting loss encourages each feature of $Z_T$ to be independent of features in $Z_A$ thus suppressing the sensitive-encoded covariances that cause the unfairness. 

\subsection{Overall Training} \label{sec:overall}
In this section, we introduce the overall training schema as shown in Fig.~\ref{fig:method}~(c). For the sensitive branch, \revision{we noticed that training a shared encoder can pose a risk of sensitive-related features being used for the target classification
}\cite{dullerud2022fairness}, and we pre-train separate $\left\{\phi_A,h_{A_1},...,h_{A_m}\right\}$ for multiple sensitive attributes using the sensitive loss as $L_{sens}=\frac{1}{m}\sum_{i\in[m]}L_{A_i}$.
Here we use cross-entropy loss as $L_{A_i}$ for the $i$-th sensitive attribute. Hence $p(z_A|x)$ and $p(a|z_A)$ in Eq.~\eqref{eq:inde} can be obtained. Then, the multi-sensitive space $\cS_A$ is constructed as in Section~\ref{sec:space} over the training data.
For the target branch, we use cross-entropy loss as our classification objective $L_T$ to supervise the training of $\phi_T$ and $h_T$ and estimate $p(z_T|x)$ and $p(y|z_T)$ in Eq.~\eqref{eq:inde} respectively. Here we do not make additional constraints to $L_T$, which means it can be replaced by any other task-specific losses. At last, we apply our column and row orthogonality losses $L_{corth}$ and $L_{rorth}$ to representations as introduced in Section~\ref{sec:space} and Section~\ref{sec:row} along with detached $\cS_A$ and $Z_A$ to approximate independence between $p(z_A|x)$ and $p(z_T|x)$. The overall target objective is given as:
\begin{align}\label{eq:targ_loss}
    L_{targ} &= L_T  + \lambda_{c}L_{corth} + \lambda_{r} L_{rorth},
\end{align}
where $\lambda_{c}$ and $\lambda_{r}$ are hyper-parameters to weigh orthogonality and balance fairness and utility.



\section{Experiments}
\begin{table}[!tp]
    \centering
    \renewcommand{\arraystretch}{1}
    \newcommand\Tstrut{\rule{0pt}{2.4ex}}       
    \caption{CheXpert dataset statistics and group positive rate $p(y=1|a)$ regarding \textit{pleural effusion} with three sensitive attributes race, sex, and age.}
    \label{tab:data}
    \scalebox{0.95}{
    \begin{tabular}{p{42pt} >{\centering\arraybackslash}p{40pt} >{\centering\arraybackslash}p{100pt} >{\centering\arraybackslash}p{74pt} >{\centering\arraybackslash}p{74pt}}
    \hline\hline
        \multirow{3}{*}{\rule{0pt}{3ex} Dataset} & \multirow{3}{*}{\rule{0pt}{3ex}\#Sample} & \multicolumn{3}{c}{Group Positive Rate } \Tstrut\\
        \cline{3-5}
         &  & Race & Sex  &\ Age\Tstrut\\ [-1pt]
         &&(White/Non-white/gap) & (Male/Female/gap) & (>60/$\leq$ 60/gap)\\ [1pt]
         \hline
         Original & 127130 & .410/.393/.017 & .405/.408/.003 & .440/.359/.081 \\ 
         Augmented   & 88215 & .264/.386/.122 & .254/.379/.125 & \ .264/.386/.122 \Tstrut\\
         \hline\hline
    \end{tabular}}
\end{table}
\subsection{Setup}\label{sec:setup}
\textbf{Dataset.} We adopt CheXpert dataset \cite{irvin2019chexpert} to predict \textit{Pleural Effusion} in chest X-rays, as it's crucial for chronic obstructive pulmonary disease diagnosis with high incidence.
Subgroups are defined based on the following binarized sensitive attributes: \textit{self-reported race} and \textit{ethnicity}, \textit{sex}, and \textit{age}. Note that data bias (positive rate gap) is insignificant in the original dataset (see Table~\ref{tab:data}, row 'original').  To demonstrate the effectiveness of bias mitigation methods, we amplify the data bias by (1) firstly dividing the data into different groups according to the conjunction of multi-sensitive labels; (2) secondly calculating the positive rate of each subgroup; (3) sampling out patients and increase each subgroup's positive rate gap to 0.12 (see Table~\ref{tab:data}, row `augmented'). 
We resize all images to $224\times224$ and split the dataset into a 15\% test set, and an 85\% 5-fold cross-validation set.

\noindent\textbf{Evaluation metrics.} We use the area under the ROC curve (AUC) to evaluate the utility of classifiers. To measure fairness, we follow \cite{roh2020fairbatch} and compute subgroup disparity with respect to ED (denoted as $\disp_\mathrm{ED}$, which is based on true positive rate (TPR) and true negative rate (TNR)) in Eq. \eqref{eq:cjme} as:
\begin{equation} \label{eq:fair-metric}
    \mathrm{\Delta}_\mathrm{ED}=\max_{y\in\mathcal{Y},\pi_1,\pi_2 \in \mathcal{A}}\left|P(\hat{Y}=y|A=\pi_1,Y=y)-P(\hat{Y}=y|A=\pi_2,Y=y)\right|.
\end{equation}
We also follow \cite{zhang2022improving} and compare subgroup disparity regarding AUC (denoted as $\disp_\mathrm{AUC}$), which gives a threshold-free fairness metric. Note that we evaluate disparities both \textit{jointly} and \textit{individually}. The \textit{joint} disparities \revision{assess multi-sensitive fairness by computing disparity across subgroups defined by the combination of multiple sensitive attributes $\mathcal{A}$}. \revision{On the other hand,} \textit{individual} disparities \revision{are calculated based on} a specific binary sensitive attribute $A_i$. 

\begin{table}[!tp]
\centering
\renewcommand{\arraystretch}{1}
\newcommand\Tstrut{\rule{0pt}{2.6ex}}         
\newcommand\Bstrut{\rule[3ex]{0pt}{0pt}} 
\newcommand\std{\fontsize{8}{9}\selectfont}

\caption{Comparasion of predicting \textit{Pleural Effusion} on CheXpert dataset. We report the mean and standard deviation of 5-fold models \textbf{trained with multi-sensitive attributes}. AUC is used as the utility metric, and fairness is evaluated using disparities among subgroups defined on multi-sensitive attributes \textit{\{Race, Sex, and Age\} jointly} and \revision{each of the attribute \textit{individually}}.}
\label{tab:comparison}
\noindent\makebox[7cm]{
\begin{tabular}{
>{\arraybackslash}p{55pt}|
>{\centering\arraybackslash}p{35pt}
*4{|*{2}{>{\centering\arraybackslash}p{28pt}}}}
\hline\hline \multirow{3}{*}{\rule{0pt}{4ex}Methods} & \multirow{3}{*}{\rule{0pt}{4ex}AUC ($\uparrow$)} & \multicolumn{8}{c}{Subgroup Disparity ($\downarrow$)} \Tstrut\\[2pt]
\cline{3-10}
 && \multicolumn{2}{c}{Joint}\vrule & \multicolumn{2}{c}{Race}\vrule & \multicolumn{2}{c}{Sex}\vrule & \multicolumn{2}{c}{Age} \Tstrut\\[1pt]
 \cline{3-10}
 && $\disp_\mathrm{AUC}$ & $\disp_\mathrm{ED}$ & $\disp_\mathrm{AUC}$ & $\disp_\mathrm{ED}$ & $\disp_\mathrm{AUC}$ & $\disp_\mathrm{ED}$ & $\disp_\mathrm{AUC}$ & \ $\disp_\mathrm{ED}$  \Tstrut\\
 \hline\hline
 \multirow{2}{*}{ERM \cite{ERM}} & 0.863 & 0.119 & 0.224 & 0.018 & 0.055 & 0.046 & 0.142 & 0.023 & \ 0.038 \Tstrut\\ [-3pt]
  & \std(.005) & \std(.017) & \std(.013) & \std(.009) & \std(.017) & \std(.008) & \std(.014) & \std(.004) & \std(.010) \\ [1pt]
 \hline
 \multirow{2}{*}{G-DRO \cite{sagawa2019distributionally}} & 0.854 & 0.101 & 0.187 & 0.015 & 0.048 & 0.034 & 0.105 & 0.035 & \ 0.051  \Tstrut\\ [-3pt]
 & \std(.004) & \std(.012) & \std(.034) & \std(.003) & \std(.014) & \std(.010) & \std(.025) & \std(.002) & \std(.010)\\ [1pt]
 \multirow{2}{*}{JTT \cite{liu2021just}} & 0.834 & 0.103 & 0.166 & 0.019 & 0.056 & 0.026 & 0.079 & 0.017 & 0.030 \\ [-3pt]
 & \std(.020) & \std(.017) & \std(.023) & \std(.008) & \std(.016) & \std(.002) & \std(.004) & \std(.006) & \std(.007)\\ [1pt]
 \multirow{2}{*}{Adv \cite{wadsworth2018achieving}} & 0.854 & 0.089 & 0.130 & 0.017 & \textbf{0.027} & 0.022 & 0.039  & 0.016 & 0.023 \\ [-3pt]
 & \std(.002) & \std(.009) & \std(.018) & \std(.004) & \std(\bf .009) & \std(.003) & \std(.008) & \std(.004) & \std(.004)\\ [1pt]
 \multirow{2}{*}{BR-Net \cite{adeli2021representation}} & 0.849 & 0.113 & 0.200 & 0.018 & 0.051 & 0.037 & 0.109 & 0.027 & 0.039   \\ [-3pt]
 & \std(.001) & \std(.025) & \std(.023) & \std(.008) & \std(.013) & \std(.012) & \std(.025) & \std(.006) & \std(.006)\\ [1pt]
 \multirow{2}{*}{PARADE \cite{dullerud2022fairness}} & 0.857 & 0.103 & 0.193 & 0.017 & 0.052 & 0.042 & 0.104 & 0.026 & 0.031 \\ [-3pt]
 & \std(.002) & \std(.022) & \std(.032) & \std(.002) & \std(.010) & \std(.006) & \std(.023) & \std(.006) & \std(.011)\\ [1pt]
 \multirow{2}{*}{Orth \cite{sarhan2020fairness}} & 0.856  & 0.084 & 0.177  & 0.012 & 0.045 & 0.022 & 0.083 &  0.025 & 0.032 \\   [-3pt]
 & \std(.007) & \std(.022) & \std(.016) & \std(.005)& \std(.012) &  \std(.009) & \std(.012)&\std(.006)&\std(.005)\\ [1pt]
 \hline 
  \multirow{2}{*}{FCRO (ours)} & \ \textbf{0.858} \Tstrut & \textbf{0.057} & \textbf{0.107} & \textbf{0.012} & 0.033 & \textbf{0.015} & \textbf{0.024} & \textbf{0.013} & \textbf{0.019}\\ [-3pt]
 & \std(\bf .001) & \std(\bf .022) & \std(\bf .013) & \std(\bf .003) & \std(.008) & \std(\bfseries .004) & \std(\bf .008) & \std(\bf .004) & \std(\bf .006)\\ [1pt]

\hline\hline
\end{tabular}}
\label{performance}
\end{table}
\noindent\textbf{Implementation details.} In our implementation, all methods use the same training protocol. We choose DenseNet-121 \cite{huang2017densely} as the backbone, but replace the final layer with a linear layer to extract 128-dimensional representations. The optimizer is Adam with learning rate of $1e^{-4}$, and weight decay of $4e^{-4}$. We train for 40 epochs with a batch size of 128. We sweep a range of hyper-parameters for each method and empirically set $\lambda_c=80$, $\lambda_r=500$, and $k=3$ for \ours{}. We train models in 5-fold with different random seeds. In each fold, we sort all the validations according to utility and select the best model with the lowest average $\disp_\mathrm{ED}$ regarding each sensitive attribute among the top 5 utilities. 

\noindent\textbf{Baselines.} We compare our method with (\romannumeral1) G-DRO \cite{sagawa2019distributionally} and (\romannumeral2) JTT \cite{liu2021just} -- methods that seek low worst-group error by minimax optimization on group fairness and target task error, which can be naturally regarded as multi-sensitive fairness methods by defining subgroups with multi-sensitive attributes conjunctions. We also 
extend recent state-of-the-art fair representation learning methods on single sensitive attributes to multiple ones and compare our method with them, including (\romannumeral3) Adv \cite{wadsworth2018achieving} and (\romannumeral4) BR-Net \cite{adeli2021representation} -- methods that achieve fair representation via disentanglement using adversarial training, (\romannumeral5) PARADE \cite{dullerud2022fairness} -- a state-of-the-art method that adversarially eliminates mutual information between target and sensitive attribute representations 
and (\romannumeral6) Orth~\cite{sarhan2020fairness} \revision{-- a recent work closest to our method}. Orth hard codes the means of both sensitive and target prior distributions to orthogonal means and re-parameterize the encoder output on the orthogonal priors. 
Besides, we give the result of (\romannumeral7) ERM \cite{ERM} -- the vanilla classifier trained without any bias mitigation techniques.
\begin{figure}[!tp]
    \centering
\includegraphics[width=0.8\linewidth]{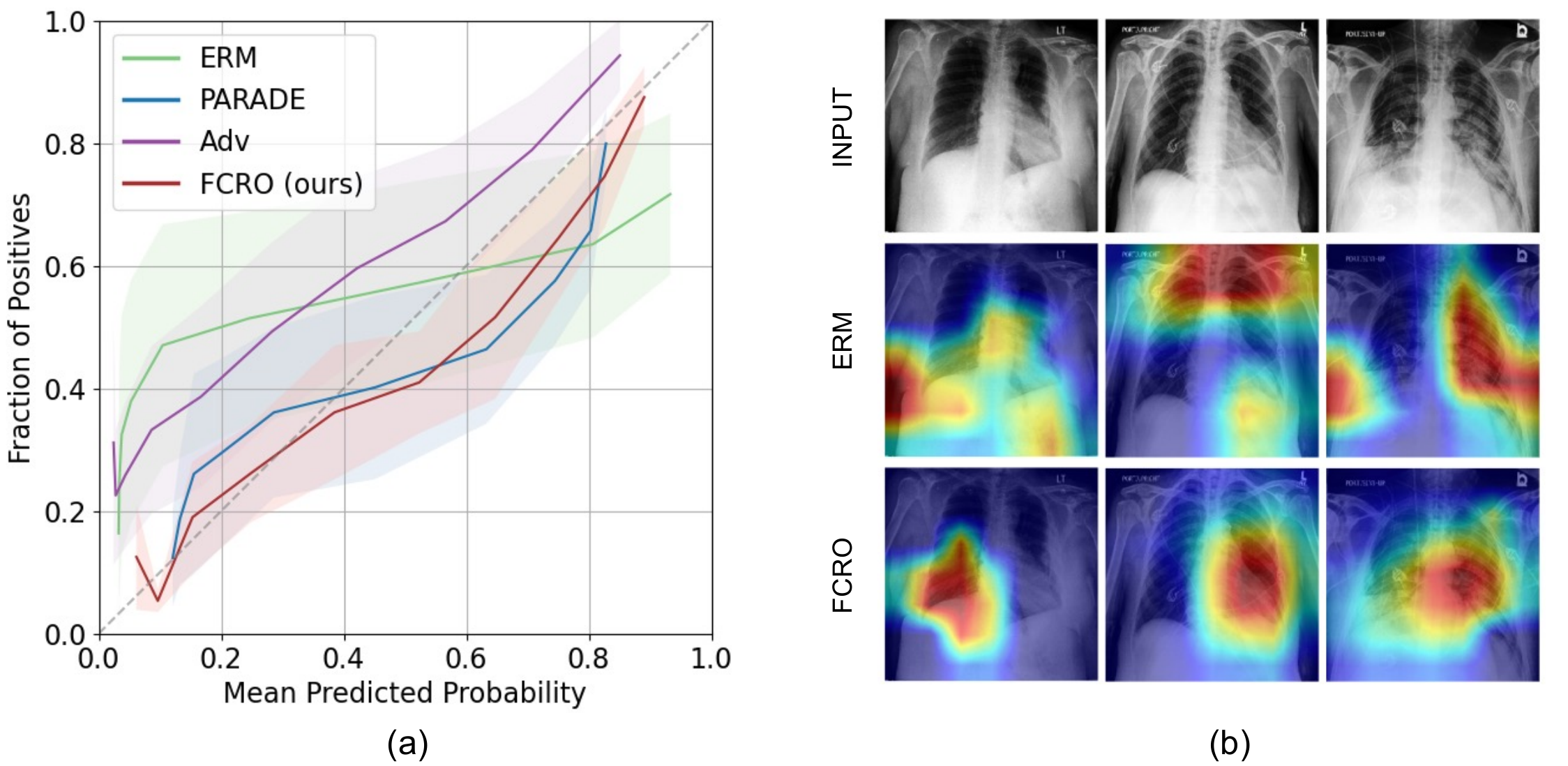}
    \caption{(a) Subgroup calibration curves. We report the mean (the line) and standard deviation (the shadow around it) of different subgroups defined by the conjunction of race, sex, and age. Larger shadow areas represent more severe unfairness. (b) Class activation map \cite{chattopadhay2018grad} generated from vanilla ERM \cite{ERM} and \ours{} (ours).}
    \label{fig:activation}
\end{figure}
\subsection{Comparsion with Baselines}
\textbf{Quantitative results.} 
We summarize quantitative comparisons in Table~\ref{tab:comparison}. It can be observed that all the bias mitigation methods can improve fairness compared to ERM~\cite{ERM} at the cost of utility. While ensuring considerable classification accuracy, \ours{} achieves significant fairness improvement both \textit{jointly} and \textit{individually}, demonstrating the effectiveness of our representation orthogonality motivation. To summarize, compared with the best performance in each metric, \ours{} reduced classification disparity on subgroups with \textit{joint} $\disp_\mathrm{AUC}$ by 2.7\% and \textit{joint} $\disp_\mathrm{ED}$ by 2.3\% respectively, and experienced 0.5\% $\disp_\mathrm{AUC}$ and 0.4\% $\disp_\mathrm{ED}$ boosts regarding the average of three sensitive attributes. 
As medical applications are sensitive to classification thresholds, we further show calibration curves with the mean and standard deviation of subgroups defined on the conjunction of multiple sensitive attributes in Fig.~\ref{fig:activation}~(a). The vanilla ERM \cite{ERM} suffers from biased calibration among subgroups. Fairness algorithms can help mitigate this, while \ours{} shows the most harmonious deviation and the most trustworthy classification.
\noindent\textbf{Qualitative results.} 
We present the class activation map~\cite{chattopadhay2018grad} in Fig.~\ref{fig:activation}~(b). We observe that the vanilla ERM \cite{ERM} model tends to look for sensitive evidence outside the lung regions, \eg breast, which threatens unfairness. \ours{} focuses on the pathology-related part only for fair \textit{Pleural Effusion} classification, which visually confirms the validity of our method. 
\begin{figure}[!tp]
    \centering
    \includegraphics[width=0.98\linewidth]{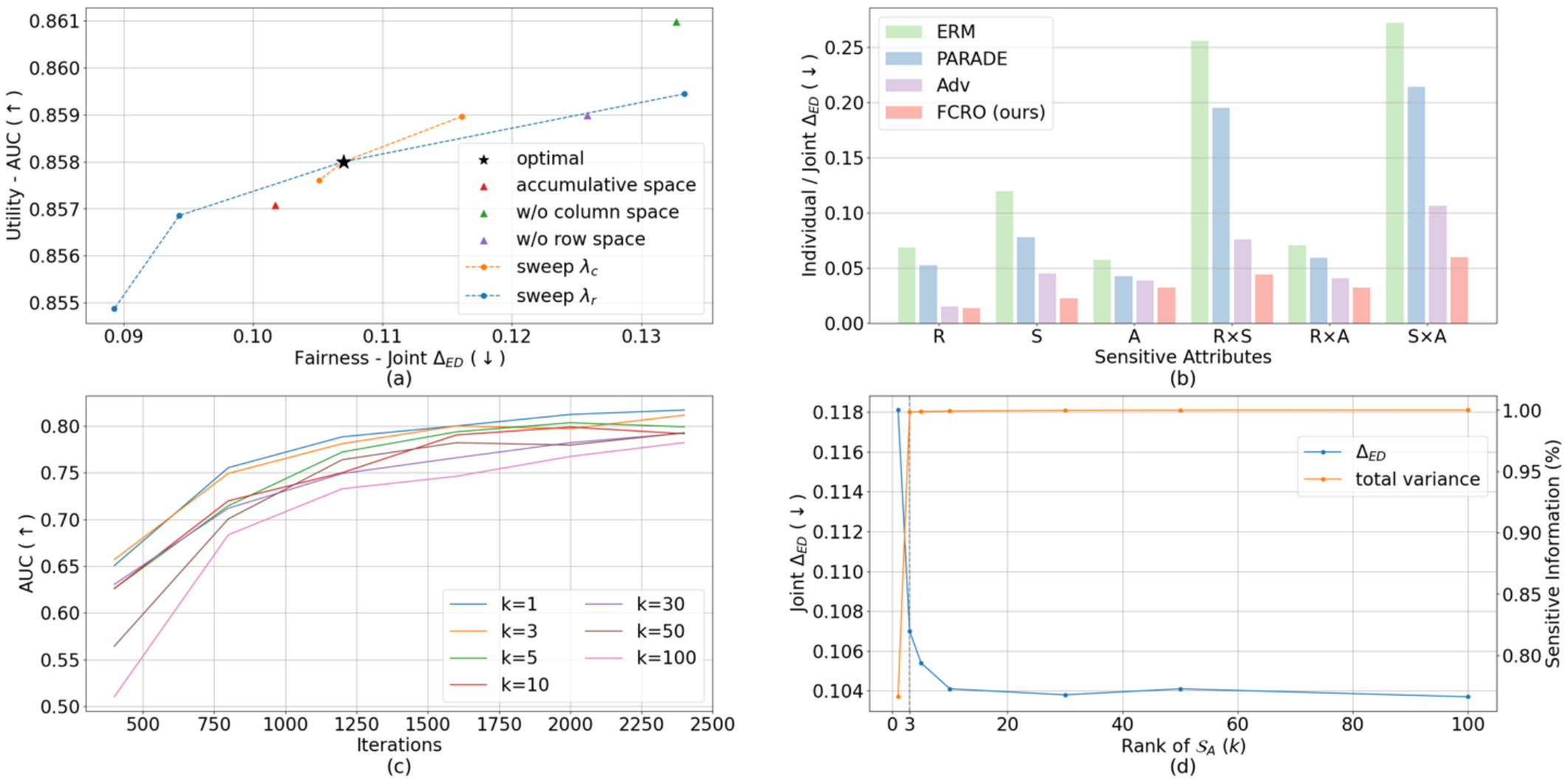}
    \caption{(a) Fairness-utility trade-off. The perfect point lies in the top left corner. We report ablations and Pareto fronts of the sweep of hyperparameters. (b) Fairness when \textbf{trained with various combinations} of three sensitive attributes: race (R), sex (S), and age (A). (c) AUC convergence with different rank $k$ of $\cS_A$. (d) Fairness and total variance (the percentage of sensitive information captured by $\cS_A$) under different $k$.}
    \label{fig:curve}
\end{figure}

\subsection{Ablation Studies}

\noindent\textbf{Loss modules and hyperparameters.} 
\label{module_ablation}
\revision{To evaluate the impact of loss weights and critical components of \ours{}, we delve into different choices of hyperparameters and alternative designs.} As depicted in Fig.~\ref{fig:curve}~(a), \revision{we showcase the significance of the key components and the Pareto fronts (\ie the set of optimal points) curve that results from a range of hyperparameters }$\lambda_c$ and $\lambda_r$. \revision{Increasing $\lambda_c$ and $\lambda_r$ resulted in a small decrease in AUC (0.5\% and 0.12\%, respectively). However, this was offset by significant gains in fairness (4\% and 1.5\%, respectively). The observation confirms the goal of managing multiple sensitive attributes effectively without sacrificing accuracy.} Besides,  We observe that removing either column or row space orthogonality shows a decrease in \textit{joint} $\disp_\mathrm{ED}$ of 2.4\% and 1.8\% respectively, but still being competitive. We also observe accumulative space introduced in Section \ref{sec:space} achieves a comparable performance.

\noindent\textbf{Training with different sensitive attributes.} We present an in-depth ablation study on multiple sensitive attributes in Fig.~\ref{fig:curve}~(b), where models are trained with various numbers and permutations of attributes. We show all methods perform reasonably better than ERM when trained with a single sensitive attribute but \ours{} brought significantly more benefits when trained with the union of discriminated attributes (e.g., Sex $\times$ Age), which consolidate \ours{}'s ability for multi-sensitive attributes fairness. \ours{} stand out among all methods.

\noindent\textbf{Different rank $k$ for $\wt{\cS}_A$.} 
We show the effect of choosing different $k$ for column space orthogonality. As shown in Fig.~\ref{fig:curve}~(c), a lower rank $k$ benefits convergence of the model thus improving accuracy, which validates our insights in Section.~\ref{sec:space} that lower sensitive space rank will improve the utility of target representations. In Fig.~\ref{fig:curve} (d), we show that $k=3$ is enough to capture over $95\%$ sensitive information and keep increasing it does not bring too much benefit for fairness, thus we choose $k=3$ to achieve the best utility-fairness trade off.

\section{Conclusion and Future Work} 
This work studies an essential yet under-explored fairness problem in medical image classification where samples are with sets of sensitive attributes. We formulate this problem mathematically and propose a novel fair representation learning algorithm named \ours{}, which pursues orthogonality between sensitive  and target representations. Extensive experiments on a large public chest X-rays demonstrate that \ours{} significantly boosts the fairness-utility trade-off both \textit{jointly} and \textit{individually}. Moreover, we show that \ours{} performs stably under different situations with in-depth ablation studies. For future work, we plan to test the scalability of \ours{} on an extremely large number of sensitive attributes. 
 \vspace*{-0.8\baselineskip}
\section*{Acknowledgments}
 \vspace*{-0.8\baselineskip}
This work is supported in part by the Natural Sciences and Engineering Research Council of Canada (NSERC), Public Safety Canada (NS-5001-22170), in part by NVIDIA Hardware Award, and in part by the Hong Kong Innovation and Technology Commission under Project No. ITS/238/21.

%
%
%
\bibliographystyle{splncs04}
 \vspace*{-1\baselineskip}
\bibliography{references}


\end{document}